\pdfoutput=1
\documentclass[11pt]{article}

\usepackage{graphicx}
\usepackage{amsmath}
\usepackage{amssymb}
\usepackage{float}
\usepackage{caption}
\usepackage[authoryear]{natbib}
\usepackage[colorlinks=true, allcolors=blue]{hyperref}

\title{Vision Transformer for Transient Noise Classification}

\author{
D. Srivastava$^{1}$ \and A. Niedzielski$^{1}$\\[4pt]
\small $^{1}$Institute of Astronomy, Nicolaus Copernicus University in Toruń,\\
\small ul. Gagarina 11, 87-100 Toruń, Poland
}
\date{}

\begin{document}
\maketitle

\begin{abstract}
Transient noise (glitches) in LIGO data hinders the detection of gravitational waves (GW). The Gravity Spy project has categorised these noise events into various classes. With the O3 run, there is the inclusion of two additional noise classes and thus a need to train new models for effective classification.

\textbf{Aims:} We aim to classify glitches in LIGO data into 22 existing classes from the first run plus 2 additional noise classes from O3a using the Vision Transformer (ViT) model.

\textbf{Methods:} We train a pretrained Vision Transformer (ViT-B/32) model on a combined dataset consisting of the Gravity Spy dataset with the additional two classes from the LIGO O3a run.

\textbf{Results:} We achieve a classification efficiency of 92.26\%, demonstrating the potential of Vision Transformer to improve the accuracy of gravitational wave detection by effectively distinguishing transient noise.\\
\vspace{15pt}
\end{abstract}

\section{Introduction}

Transient noise (glitches) are of many types, some with known environmental or instrumental sources and some whose causes are unknown. They hinder the detection and analysis of gravitational wave signals significantly. Therefore identifying and classifying glitches is important to improve the gravitational wave signals' quality.

Previous approaches for classification have used convolutional neural networks (CNNs) such as VGG16 and ResNet50 (\cite{George:2017qtr}). CNNs hierarchically perform feature extraction. These models have achieved considerable success, but they have limitations, particularly in capturing long-range dependencies within images and dealing with the high and low variability in glitch patterns.

Vision Transformers (ViTs), introduced by \cite{dosovitskiy2021imageworth16x16words}, offers a new approach by applying transformer models directly to sequences of image patches. They use self-attention mechanisms to find any overall dependency and connections between image patches directly. This method has shown remarkable performance on image classification tasks without inductive biases. Recently, transformers have also found applications in various areas of astronomy, as reviewed by \cite{tanoglidis2023transformersscientificdatapedagogical}. In works like stellar classification (\cite{universe10050214}) and the identification of asteroids interacting with secular
resonances (\cite{CARRUBA2025116346}), Vision Transformers have already performed better than most CNNs.

In this work, we use a pre-trained Vision Transformer (ViT-B/32) model (\cite{NEURIPS2019_9015}) to classify glitches in the first Gravity Spy dataset (\cite{coughlin2018updated}) with the two additional noise classes added from the O3a run.

\section{Methods}
In this section, we will provide a brief description of the dataset we used and the process of training the ViT-B/32 model.

\subsection{Dataset}

Gravity Spy is a citizen project that operates through the Zooniverse platform. It involves public volunteers to classify images of glitches into pre-identified morphological categories. Gravity Spy categorises all noise events detected by the Omicron trigger pipeline (\cite{ROBINET2020100620}). It identifies signals with a signal-to-noise ratio above 7.5 and peak frequencies between 10-2048 Hz. The Gravity Spy dataset included 22 classes for O1 and O2 data.

For our work, we extended the dataset to include two additional noise classes from the O3a run: Blip\_Low\_Frequency and Fast\_Scattering. These new classes were prepared from the LIGO O3a data (\cite{Glanzer_2023}) using the \texttt{gwpy} library (\cite{gwpy}) to fetch gravitational wave data segments around event times. These segments were then transformed into spectrograms using Q-transform and saved as images\footnote{\href{https://www.kaggle.com/datasets/divyansh050/ligo-noise-classifications-livingston-o3a}{LIGO O3 Noise Classifications}}.

For the two additional noise classes as well, we followed a 7:1.5:1.5 split for training, validation, and test set as in the Gravity Spy dataset. To reduce class imbalance, 3334 images for each of the two new classes from the O3a run were selected and split in the given ratio.

\begin{figure}[h]
    \centering
    \includegraphics[width=.85\linewidth]{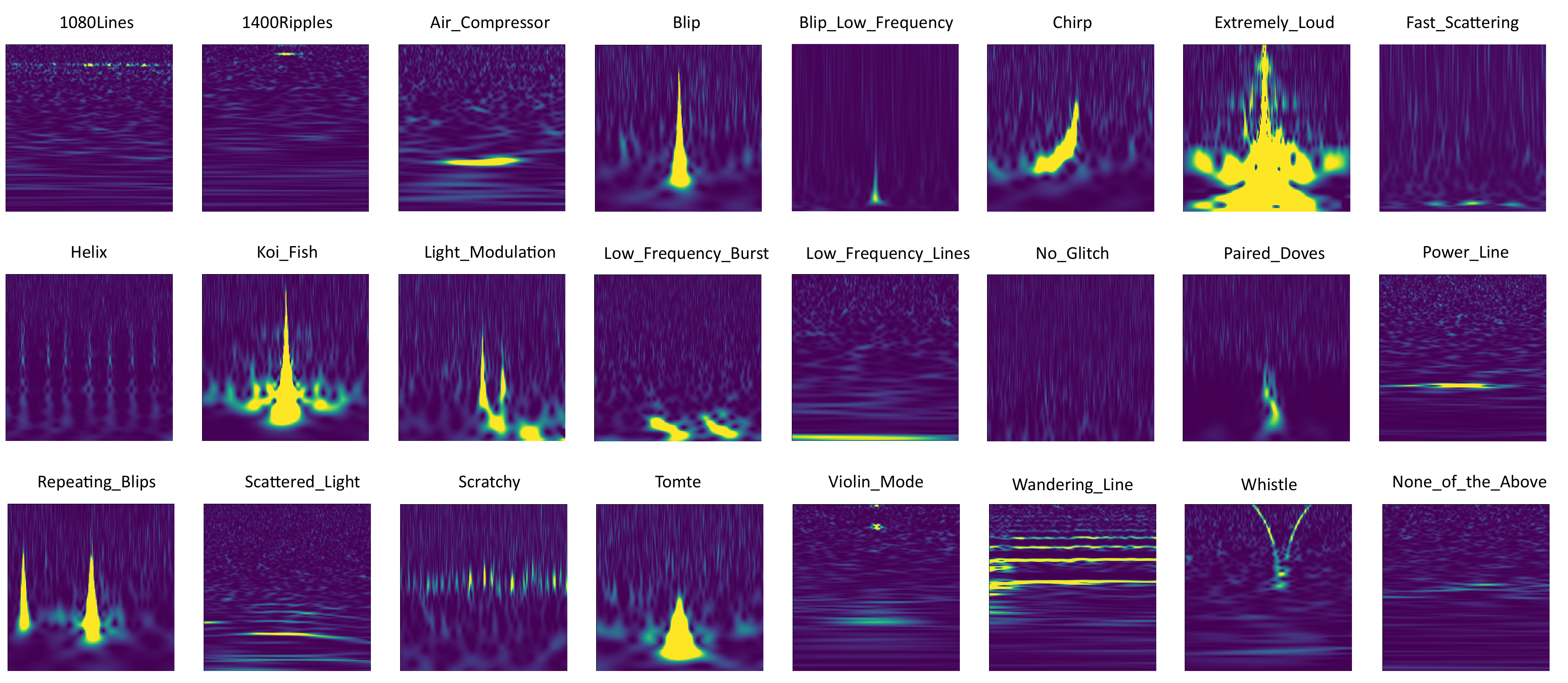}
    \caption{Spectrograms showing different glitch classes from the first Gravity Spy Dataset and additional classes Blip\_Low\_Frequency, Fast\_Scattering from O3a. The spectrograms for Blip\_Low\_Frequency and Fast\_Scattering were \(\pm\) 0.5 seconds around the events; zoomed in for identifiability.}
    \label{fig1}
\end{figure}

\subsection{Training}

The Vision Transformer (ViT) model processes an image by reshaping it into a sequence of flattened 2D patches. Specifically, an image \( x \in \mathbb{R}^{H \times W \times C} \) is converted into a sequence of patches \( x_p \in \mathbb{R}^{N \times (P^2 \cdot C)} \). Here, \((H, W)\) are the dimensions of the original image, \(C\) represents the number of channels, \((P, P)\) is the resolution of each image patch, and \(N = \frac{HW}{P^2}\) is the number of patches.

\begin{figure}[h]
    \centering
    \includegraphics[width=.7\linewidth]{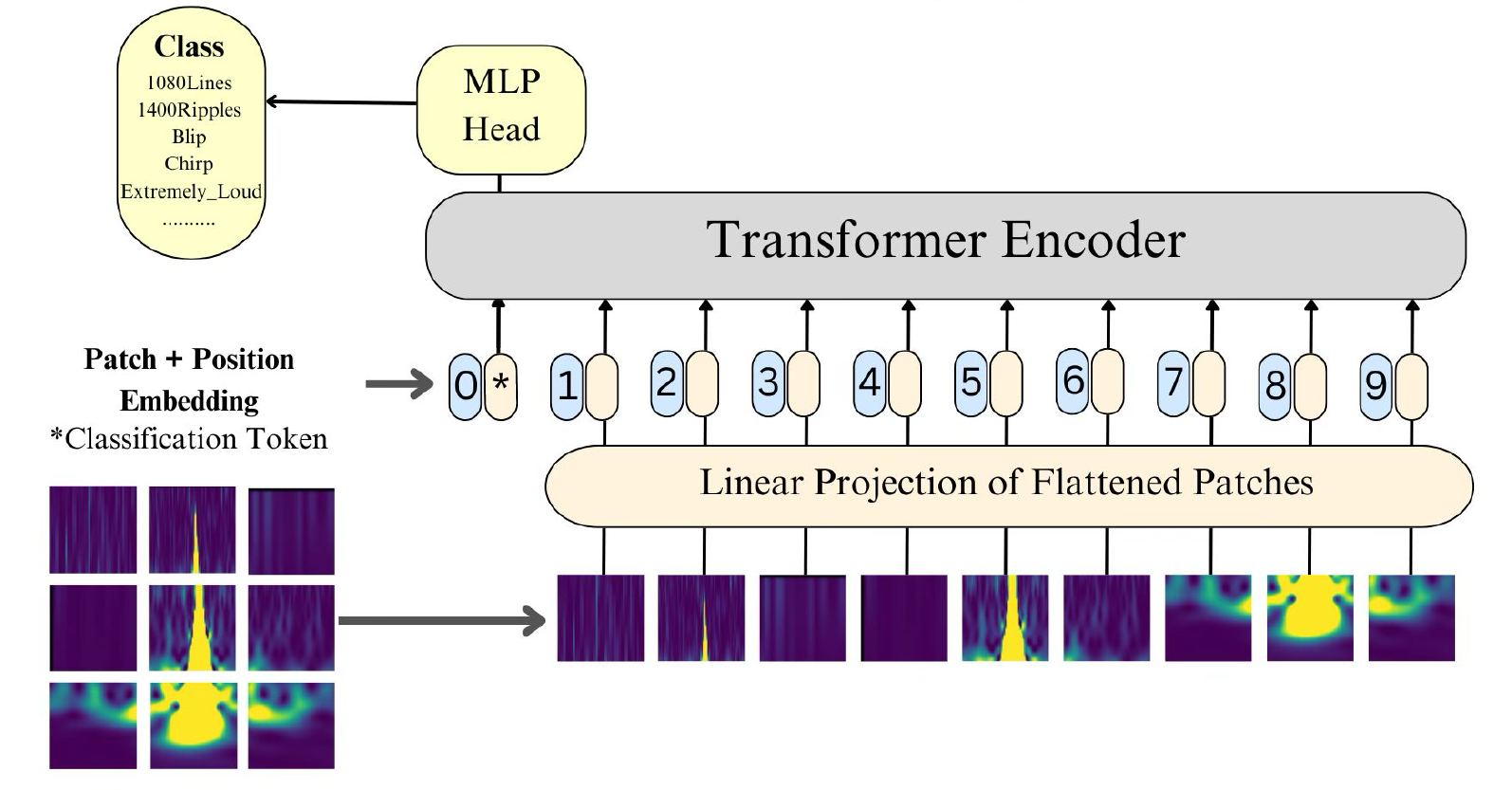}
    \caption{The image is divided into patches, each of which is linearly embedded. Positional embeddings are added, and the resulting sequence of vectors is fed into a standard Transformer encoder. For classification, an additional classification token is appended to the sequence. The MLP head then classifies the output into one of 24 classes. Illustration is inspired by \cite{dosovitskiy2021imageworth16x16words}.}
    \label{fig2}
\end{figure}

For our specific case, the images were resized to the dimensions of \((224, 224)\), the number of channels \(C\) is \(3\) (for RGB images), and the patch resolution \((P, P)\) is \((32, 32)\) because we are using the ViT-B/32 model. Thus, the number of patches \(N\) is \( \frac{224 \times 224}{32 \times 32} = 49 \). We illustrate this in Figure \ref{fig2} with 3 x 3 patches instead of 7 x 7 for simplicity. These flattened patches are linearly projected into a lower-dimensional space, creating patch embeddings. Positional encodings are added to the patch embeddings to retain the spatial information. These special embeddings are then passed through multiple layers of the Transformer encoder, which includes the multi-head self-attention mechanism and feed-forward neural networks. Each encoder layer processes the input embeddings, captures the relationships between patches, and refines the feature representations.

To train the model, we first created a combined dataset by adding the two new classes from the O3a run to the first Gravity Spy dataset. Images were resized, converted to tensors, and normalised. Data loaders for the training, validation, and test datasets were then created with a batch size of 32. The Adam optimizer with a learning rate of 0.001 was chosen to update the model's parameters during training, mainly due to its efficiency in handling sparse gradients (when only a subset of parameters receive significant updates in each iteration). It is a common choice in many deep-learning applications.

The cross-entropy loss function was used to measure the performance of the classification model, as it is appropriate for multi-class classification tasks. The cross-entropy loss \( L \) is defined as:
\[
L = -\sum_{i=1}^{n} y_i \log(\hat{y}_i)
\]
where \( y_i \) is the true label, \( \hat{y}_i \) is the predicted probability for class \( i \), and \( n \) is the number of classes. This function provides a probability value between 0 and 1 for each class.\\

The ViT-B/32 model used in this work is pre-trained on the ImageNet-1K dataset, achieving an accuracy of 75.912\% (top-1) and 92.466\% (top-5) on ImageNet-1K.\footnote{\href{https://www.kaggle.com/code/divyansh050/vitb32-ligo/notebook}{ViT-B/32 model}} All parameters were frozen except for the classifier head. The classifier head, implemented as a Multi-Layer Perceptron (MLP) with a two-layer structure and Gaussian Error Linear Unit (GELU) activation, was modified to fit the 24 classes in the combined dataset. Freezing the pre-trained layers helped retain the learned features from the ImageNet dataset. This allows the model to use this prior knowledge while training the new classifier head to specialise in our specific glitch classification task.

\begin{figure}[h]
    \centering
    \includegraphics[width=.4\textwidth]{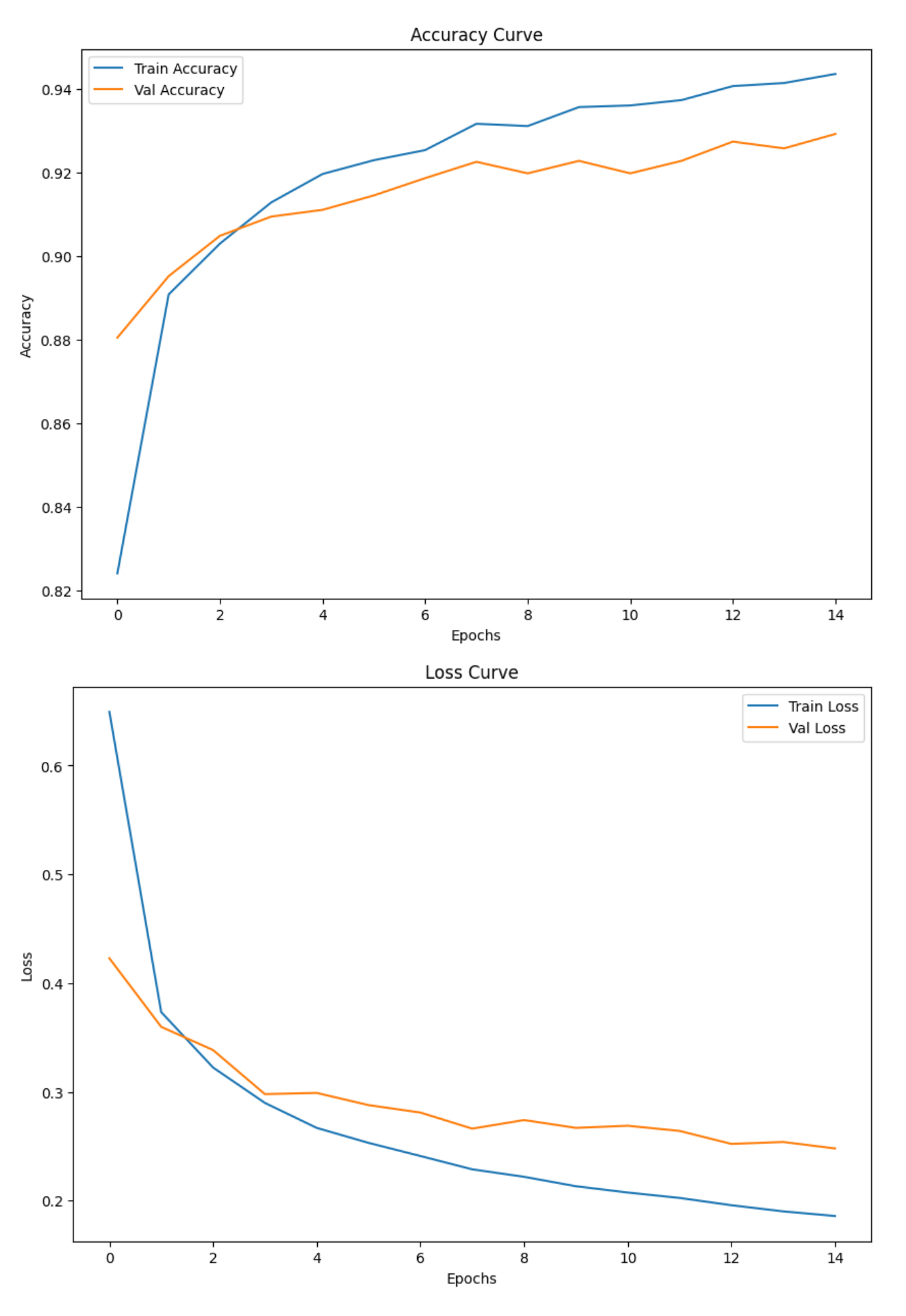}
    \captionsetup{font=small}
    \caption{Training and validation accuracy and loss over 15 epochs.}
    \label{fig:training_curves}
\end{figure}

We trained the model for 15 epochs. In each epoch, the model was trained and validated on the respective sets. The training and validation loss and accuracy were recorded to monitor the model's performance and detect any signs of overfitting. For reproducibility, random seeds were set, and the parameters of the Vision Transformer layers were frozen during training, so only the classifier head was updated. The training and validation accuracy and loss curves are shown in Figure \ref{fig:training_curves}.

During training, we observed that both the training and validation accuracy continued to improve across epochs, with minute fluctuations. The training accuracy reached over 94\%, while the validation accuracy improved to approximately 93\% in the final epoch. Similarly, both the training and validation loss decreased significantly over the 15 epochs. The difference between training and validation accuracy remained around 1--2\%. The validation loss was consistently slightly higher than the training loss and the training accuracy was marginally higher than the validation accuracy.

This difference between training and validation accuracy and loss indicates minimal overfitting which could slightly affect the model's predictions on the test data by potentially limiting its generalisation performance. After training, the best model weights based on validation accuracy were saved for future use.\footnote{\href{https://www.kaggle.com/code/divyansh050/vitb32-ligo/notebook}{Model Training Notebook}}

\newpage
\section{Results}
The performance of the Vision Transformer (ViT-B/32) model on the test dataset was evaluated using F1 Score and Accuracy. The F1 Score obtained was 92.13\%, and the Accuracy was 92.26\%.

The best-performing classes were 1080Lines, Blip, Extremely\_Loud, and Helix, all with accuracies above 98\%. On the other hand, classes like Paired\_Doves and No\_Glitch performed poorly, with accuracies of 9.09\% and 26.56\%, respectively.

In comparison, the best results using CNN architectures like VGG16 and ResNet50 have reported accuracies exceeding 98\% on the original 22-class Gravity Spy dataset (\cite{George:2017qtr}). Even pre-trained Vision Transformers typically require larger datasets to attain better performance than CNNs due to their lack of strong inductive biases (\cite{li2023deeperunderstandingretnetviewed}). While ViT-B/32 did not outperform traditional CNNs in our study, achieving over 92\% accuracy demonstrates that Vision Transformers are capable of effectively classifying transient noise in gravitational wave data. Our future work will explore unfreezing the Vision Transformer's encoder layers to potentially increase classification accuracy.

\vfill

\begin{figure}[H]
    \centering
    \includegraphics[width=.85\textwidth]{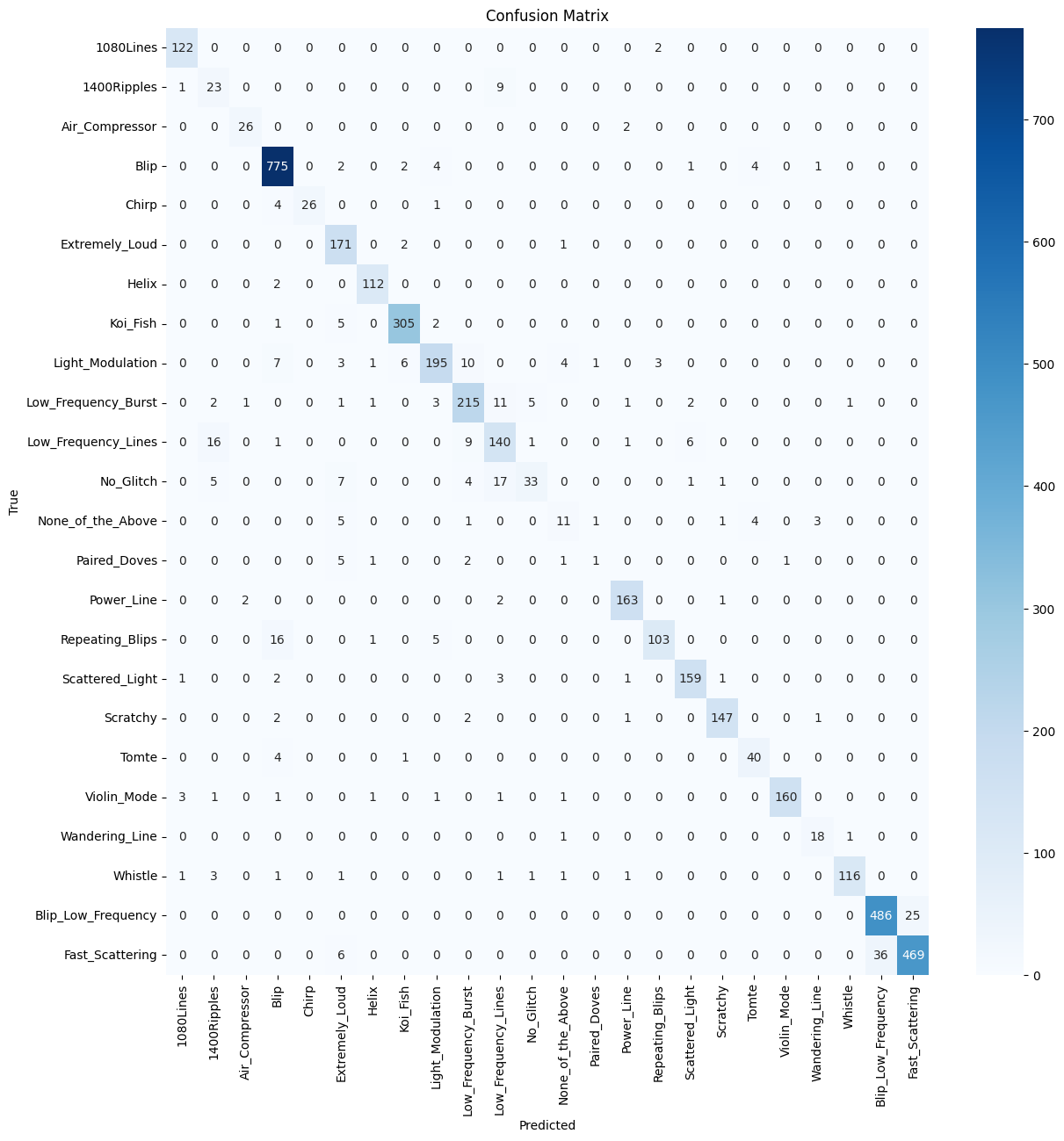}
    \captionsetup{font=small}
    \caption{Confusion Matrix for the test dataset showing the classification performance of the Vision Transformer (ViT-B/32) model across different classes.}
    \label{fig:confusion_matrix}
\end{figure}

\bibliography{references}
\bibliographystyle{plainnat}
\end{document}